\relax
\documentclass[letterpaper]{article} % DO NOT CHANGE THIS
\usepackage{jmlr2e}
\usepackage{times}  % DO NOT CHANGE THIS
\usepackage{helvet} % DO NOT CHANGE THIS
\usepackage{courier}  % DO NOT CHANGE THIS
\usepackage{url}  % DO NOT CHANGE THIS
\usepackage{graphicx,amsmath,amssymb} % DO NOT CHANGE THIS
\usepackage{natbib}  % DO NOT CHANGE THIS AND DO NOT ADD ANY OPTIONS TO IT
\usepackage{caption} % DO NOT CHANGE THIS AND DO NOT ADD ANY OPTIONS
\usepackage{subfig}

\usepackage{amsmath}
\usepackage[switch]{lineno}

\usepackage{color} % DO NOT CHANGE THIS
\newcommand{\name}{PRoFILE}

\title{Accurate and Robust Feature Importance Estimation under Distribution Shifts}
\author{
	\name Jayaraman J. Thiagarajan \email jjayaram@llnl.gov \\
	\addr Lawrence Livermore National Labs\\
	Livermore, CA,USA
	\AND
	\name Vivek Narayanaswamy \email vnaray29@asu.edu \\
	\addr Arizona State University\\
	Tempe, AZ, USA
	\AND
		\name Rushil Anirudh \email anirudh1@llnl.gov \\
	\addr Lawrence Livermore National Labs\\
	Livermore, CA,USA
	\AND
		\name Peer-Timo Bremer \email bremer5@llnl.gov \\
	\addr Lawrence Livermore National Labs\\
	Livermore, CA,USA
	\AND
	\name Andreas Spanias \email spanias@asu.edu \\
	\addr Arizona State University\\
	Tempe, AZ, USA	
}

\begin{document}

\maketitle
\begin{abstract}
With increasing reliance on the outcomes of black-box models in critical applications, post-hoc explainability tools that do not require access to the model internals are often used to enable humans understand and trust these models. In particular, we focus on the class of methods that can reveal the influence of  input features on the predicted outputs. Despite their wide-spread adoption, existing methods are known to suffer from one or more of the following challenges: computational complexities, large uncertainties and most importantly, inability to handle real-world domain shifts. In this paper, we propose PRoFILE, a novel feature importance estimation method that addresses all these challenges. Through the use of a loss estimator jointly trained with the predictive model and a causal objective, PRoFILE can accurately estimate the feature importance scores even under complex distribution shifts, without any additional re-training. To this end, we also develop learning strategies for training the loss estimator, namely contrastive and dropout calibration, and find that it can effectively detect distribution shifts. Using empirical studies on several benchmark image and non-image data, we show significant improvements over state-of-the-art approaches, both in terms of fidelity and robustness.

% A natural question of how useful the explanations are towards interpreting decisions strongly motivates the need to obtain explanations that can easily identify crucial features of data while being immune to domain shifts. Commonly adopted post-hoc strategies range from generating computationally complex, sample level explanations to scalable methods that can well approximate the black box using simpler models. Although easily interpretable, these approaches are severely challenged by increased uncertainty and inability in producing robust explanations under distribution shifts leading to inaccurate inferences. To this end, we propose XYZ, a novel feature importance estimation method that is highly accurate and robust under distribution shifts. Through the use of a loss estimator jointly optimized with the predictive model, our approach is able to generate meaningful post-hoc explanations by measuring feature level influence on the model output through the loss estimates. In addition, we propose two different learning strategies namely contrastive and dropout calibration training, that enables the estimator to identify critical features of data through the loss estimates. Our empirical studies on several benchmark image and non-image data clearly demonstrate the effectiveness of our approach over the existing solutions.    
\end{abstract}

\section{Introduction}
With the increased adoption of machine learning (ML) models in critical decision-making, post-hoc interpretability techniques are often required to enable decision-makers understand and trust these models. The \textit{black-box} nature of ML models in most real-world settings (either due to their high complexity or proprietary nature) makes it challenging to interrogate their functioning. Consequently, attribution methods, which estimate the influence of different input features on the model output, are commonly utilized to explain decisions of such black-box models. Existing approaches for attribution, or more popularly feature importance estimation, range from sensitivity analysis~\cite{ribeiro2016, lundberg2017unified}, studying change in model confidences through input feature masking~\cite{schwab2019cxplain} to constructing simpler explanation models (e.g. linear, tree- or rule-based) that mimic a black-box model~\cite{schwab2015capturing, lakkaraju2019faithful}.  

Though sensitivity analysis techniques such as LIME~\cite{ribeiro2016} and SHAP~\cite{lundberg2017unified} are routinely used to explain individual predictions of any black-box classifier, they are computationally expensive. This challenge is typically handled in practice by constructing a \textit{global} set of explanations using a submodular pick procedure~\cite{ribeiro2016}. On the other hand, despite being scalable, methods that construct simpler explanation models~\cite{lakkaraju2019faithful}
are not guaranteed to match the behavior of the original model. While the recently proposed CXPlain~\cite{schwab2019cxplain} addresses the scalability issue of feature masking methods, they are specific to the type of masking (e.g., zero masking) and the explainer needs to be re-trained if that changes (e.g., mean masking). Finally, and most importantly, it has been well documented that current approaches are highly sensitive to distribution shifts~\cite{hima2020} and vulnerable to even small perturbations.
Recently, Lakkaraju \textit{et al.}~\cite{hima2020} formalized this problem for the case of model mimicking approaches, and showed how adversarial training can be used to produce consistent explanations. In CXPlain, Schwab \textit{et al.} proposed an ensembling strategy to effectively augment explanations with uncertainty estimates to better understand the explanation quality. However, they did not study the consistency of inferred explanations under distribution shifts.

In this work, we propose PRoFILE, a novel feature importance estimation method that is highly accurate, computationally efficient and robust under distribution shifts. The key idea of our approach is to jointly train a loss estimator while building the predictive model, and generate post-hoc explanations by measuring the influence of input features on the model output using a causal objective defined on the loss estimates. Furthermore, we introduce two different learning objectives to optimize the loss estimator, namely contrastive training and dropout calibration.
Note that, once trained, the loss estimator can also be treated as a black-box. Interestingly, we find that the loss estimator is easier to train than obtaining calibrated uncertainty estimates, yet produces higher fidelity explanations.
Further, unlike existing approaches PRoFILE requires no re-training at explanation time and natively supports arbitrary masking strategies.
Finally, we show that the resulting explanations are more robust under distribution shifts and produce higher fidelity feature importances using a variety of benchmarks.
%, when compared to existing baselines, on a wide-variety of image and non-image benchmark problems.
In summary, our contributions are:
\begin{itemize}
    \item A novel feature masking-based explainability technique that is computationally efficient and is agnostic to the type of masking;
    \item Our approach is applicable to any data modality, deep architecture, or task; 
    \item Learning objectives to train a loss estimator robust to distribution shifts;

    \item Experiments on a wide variety of both synthetic and real world data demonstrating the efficacy of PRoFILE under distribution shifts. 
\end{itemize}

\section{Related Work}
Post-hoc explanation methods are the \textit{modus-operandi} in interpreting the decisions of a black box model. Broadly, these approaches can be categorized as methods that generate explanations based on (a) sensitivity analysis; (b) gradients between the output and the input features; (c) change in model confidence through input feature masking; and (d) constructing simpler explanation models that can well approximate the black box predictor. LIME~\cite{ribeiro2016} and SHAP~\cite{lundberg2017unified} are two popular sensitivity analysis methods, and they produce sample-wise, local explanations based on regression models by measuring the sensitivity of the black-box to perturbations in the input features. However, these methods are known to involve significant computational overheads. On the other hand, Saliency Maps~\cite{simonyan2013deep}, Integrated Gradients~\cite{sundararajan2017axiomatic}, Grad-CAM~\cite{selvaraju2017grad}, DeepLIFT~\cite{shrikumar17a} and a gradient based version of SHAP - DeepSHAP~\cite{lundberg2017unified}, are examples of gradient-based methods which are computationally effective. More recently, Schwab \textit{et al.} proposed CXPlain~\cite{schwab2019cxplain} and Attentive Mixture of Experts~\cite{schwab2019granger}, which are popular examples for methods that estimate model confidences through feature masking. Trained using a Granger causality-based objective~\cite{granger1969investigating}, these methods produce attention scores reflective of the feature importances, at a significantly lower computational cost. Finally, global explanation methods rely on mimicking the black-box using simpler explainer functions. For instance, ROPE~\cite{hima2020} and MUSE~\cite{lakkaraju2019faithful} construct scalable, simple linear models and decision sets, to emulate black-box models. An inherent challenge of this class of approaches is that the simple explainers are not guaranteed to match the behavior of the original model.

While these classes of methods vary in terms of their fidelity and complexity, a common limitation that has come to light recently is that explanations from most existing methods are associated with large uncertainties~\cite{zhang2019should} and are not robust under distribution shifts. Recently, Lakkaraju \textit{et al.}~\cite{hima2020} explored the use of adversarial minmax training to ensure that the mimicking explainer model is consistent with the black-box under adversarial perturbations. In contrast, we find that, without any adversarial training, PRoFILE estimates feature importances robustly under distribution shifts, is computationally scalable compared to existing local explanation methods, and produces higher fidelity explanations.

% which construct explanations by decomposing the contribution of all neurons in a network with respect to an input feature, 
\nocite{schwab2019cxplain}

\section{Proposed Approach}
In this section, we describe our approach for feature importance estimation in deep neural networks.
% In contrast to existing feature masking-style approaches, our goal is to produce explanations , in a computationally efficient manner without re-training, and producing results that are robust under distribution shifts.

\begin{figure}[t]
    \centering
    \includegraphics[width=0.9\linewidth]{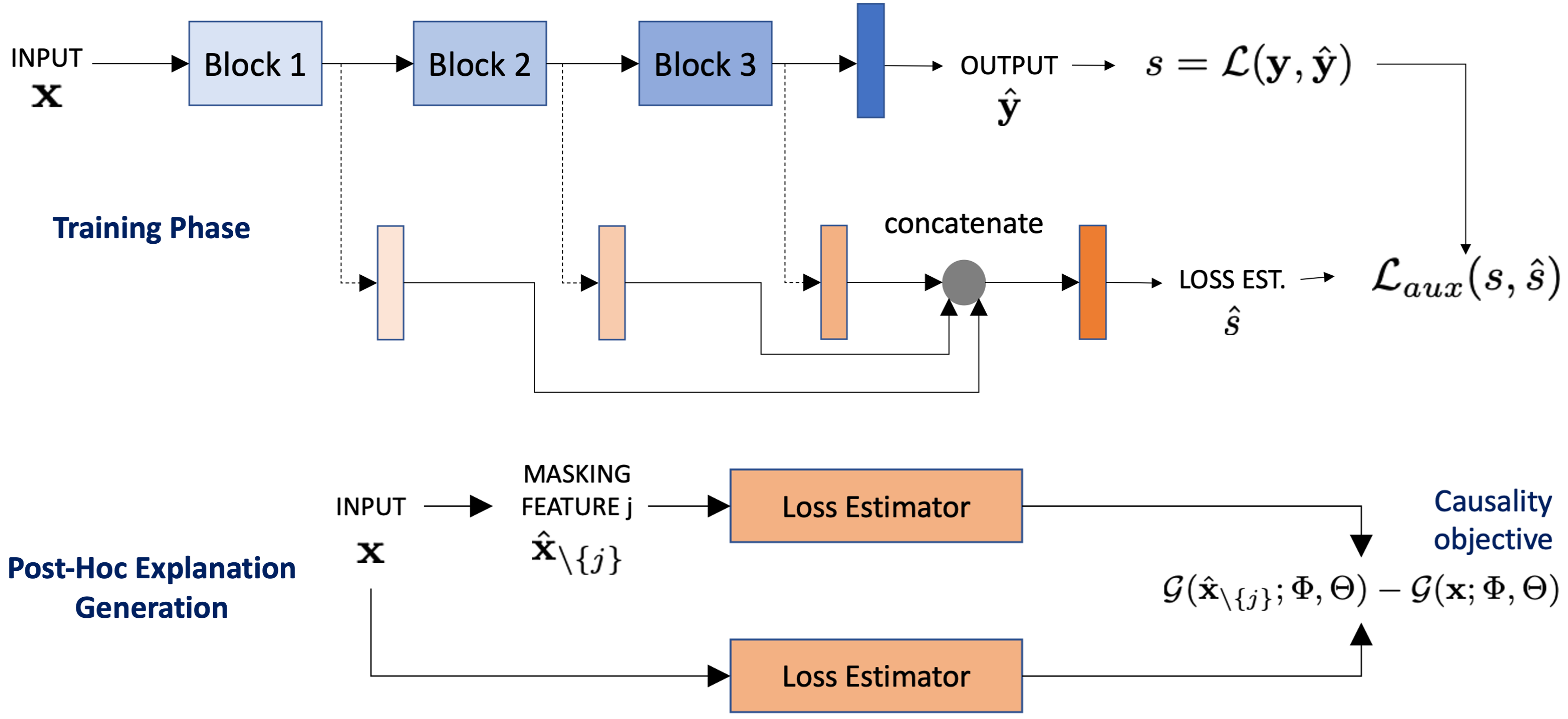}
    \caption{An illustration of the proposed approach, PRoFILE, for feature importance estimation. (top) During the training phase, we train a loss estimator along with the predictive model; (bottom) We use a Granger causality-based objective to generate post-hoc explanations using the loss estimates with no re-training.}
    \label{fig:arch}
\end{figure}

\paragraph{Predictive Model Design with Loss Estimation.}We consider the setup where we build a predictive model $\mathcal{F}(\Theta)$ which takes as input a sample $\mathbf{x} \in \mathbb{R}^d$ with $d$ features and produces the output $\hat{\mathbf{y}} \in \mathbb{R}^k$ of dimensionality $k$.
Note that this setup is deliberately unspecific in terms of both model architecture and data modality as \name{} is agnostic to either. 
Given a training set $\{(\mathbf{x}_i,\mathbf{y}_i)\}_{i=1}^N$, we optimize for the parameters $\Theta$ using the loss function $\mathcal{L}:\mathbf{y}\times \hat{\mathbf{y}} \rightarrow s$, where $s \in \mathbb{R}$. In other words, $\mathcal{L}$ measures the discrepancy between the true and predicted outputs using a pre-specified error metric. Examples include categorical cross-entropy for classification or mean-squared error for regression. 

While our approach does not need access to the training data or specifics of the training procedure while generating explanations, similar to any post-hoc interpretability method, our approach requires the training of an auxiliary network $\mathcal{G}(\Phi; \Theta)$ that takes the same input $\mathbf{x}$ and produces the output $\hat{s} \approx \mathcal{L}(\mathbf{y}, \mathcal{F}(\mathbf{x}))$.
The objective of this network is to directly estimate the fidelity for the prediction that $\mathcal{F}$ makes for $\mathbf{x}$, which we will use in order to construct our post-hoc expalantions without any additional re-training. Note that, the loss estimates implicitly provide information about the inherent uncertainties; for example, in~\cite{ash2019deep}, the gradients of loss estimates have been used to capture the model uncertainties. We define the auxiliary objective $\mathcal{L}_{aux}:s\times \hat{s} \rightarrow \mathbb{R}$, in order to train the parameters $\Phi$ of model $\mathcal{G}$. As showed in Figure~\ref{fig:arch}(top), the loss estimator $\mathcal{G}$ uses the latent representations from different stages of $\mathcal{F}$ (e.g., every layer in the case of an FCN or every convolutional block in a CNN) to estimate $\hat{s}$. We use a linear layer along with non-linear activation (ReLU in our experiments) to transform each of the latent representations from $\mathcal{F}$ and they are finally concatenated to predict the loss. During training, the gradients from both the losses are used to update the parameters $\Theta$ of model $\mathcal{F}$. 

% \ptb{I find this a bit of a tricky argument to make. You hint at the secondary network making the primary one better but I don't think we ever actually show that. At the same time I could also phrase this negatively in the sense that your explanation messes with my primary model which people might not be eager to see. Is the regularization strictly necessary or could we re-write this in terms of you could use this to regularize but you don't have to? Otherwise, I think we need some argument or experiment showing that this is actually a good thing.} 

\paragraph{Learning Objectives.}Since the proposed feature estimation strategy relies directly on the quality of the loss estimator, the choice of the loss function $\mathcal{L}_{aux}$ is crucial. In particular, our approach (see Figure \ref{fig:arch}(bottom)) is based on ranking input features using the loss values obtained by masking those features. Consequently, we expect the loss estimator to preserve the ordering of samples (based on their losses), even if the original scale is discarded. Here, we explore the use of two different objectives to minimize the discrepancy between true and estimated losses. As we will show in our empirical studies, both these objectives are highly effective at revealing the input feature importance.

\noindent \textit{(a) Contrastive Training:} This is a widely adopted strategy when relative ordering of samples needs to be preserved. Given the loss values $\{s_i, s_j\}$ for a pair of samples $\{\mathbf{x}_i, \mathbf{x}_j\}$ in a mini-batch, we adopt an objective similar to~\cite{yoo2019learning}, which ensures that the sign of the difference $(s_i - s_j)$ is preserved in the corresponding loss estimates $(\hat{s}_i - \hat{s}_j)$. Formally, we use the following contrastive loss:
\begin{align}
    \mathcal{L}_{aux}^{C} = &\sum_{(i,j)}\max \bigg(0, -\mathbb{I}(s_i,s_j) . (\hat{s}_i - \hat{s}_j) + \gamma \bigg), \\
    &\nonumber \text{where } \mathbb{I}(s_i,s_j) = \begin{cases}
1 &\text{if $s_i > s_j$},\\
0 &\text{otherwise}.
\end{cases}
    \label{eqn:laux1}
\end{align}Note, when the sign of $s_i - s_j$ is positive, we assign a non-zero penalty if the estimates $\hat{s}_j > \hat{s}_i$, i.e., there is a disagreement in the ranking of samples. Here, $\gamma$ is an optional margin hyper-parameter.

\noindent \textit{(b) Dropout Calibration:} In this formulation, we utilize prediction intervals from the model $\mathcal{F}$ and adjust the loss estimates from $\mathcal{G}$ using an interval calibration objective. The notion of interval calibration comes from the uncertainty quantification literature and is used to evaluate uncertainty estimates in continuous-valued regression problems~\cite{thiagarajan2020building}. In particular, we consider the epismetic uncertainties estimated using Monte Carlo dropout~\cite{gal2016dropout} to define the prediction interval $[\mu_{s_i} - \sigma_{s_i}, \mu_{s_i} + \sigma_{s_i}]$ for a sample $\mathbf{x}_i$. More specifically, we perform $T$ independent forward passes with $\mathcal{F}$ to compute the mean $\mu_{s_i}$ and standard deviation $\sigma_{s_i}$. For the loss estimator $\mathcal{G}$, we use the latent representations averaged across $T$ passes (for every block in Figure \ref{fig:arch}(top)) to obtain the estimate $\hat{s}_i$. Finally, we use a hinge loss objective to calibrate the estimates:
\begin{align}
    \mathcal{L}_{aux}^{DC} = \sum_i & \max\bigg(0, \hat{s}_i - (\mu_{s_i} + \sigma_{s_i}) + \xi \bigg) \\
    &+ \max\bigg(0, (\mu_{s_i} - \sigma_{s_i}) - \hat{s}_i + \xi \bigg)
\end{align}Here, $\xi$ is the optional margin parameter and the objective encourages the estimates $\hat{s}_i$ to lie in the prediction interval for $s$ from the model $\mathcal{F}$.

\begin{figure}[t]
    \centering
    \includegraphics[width=0.7\linewidth,keepaspectratio]{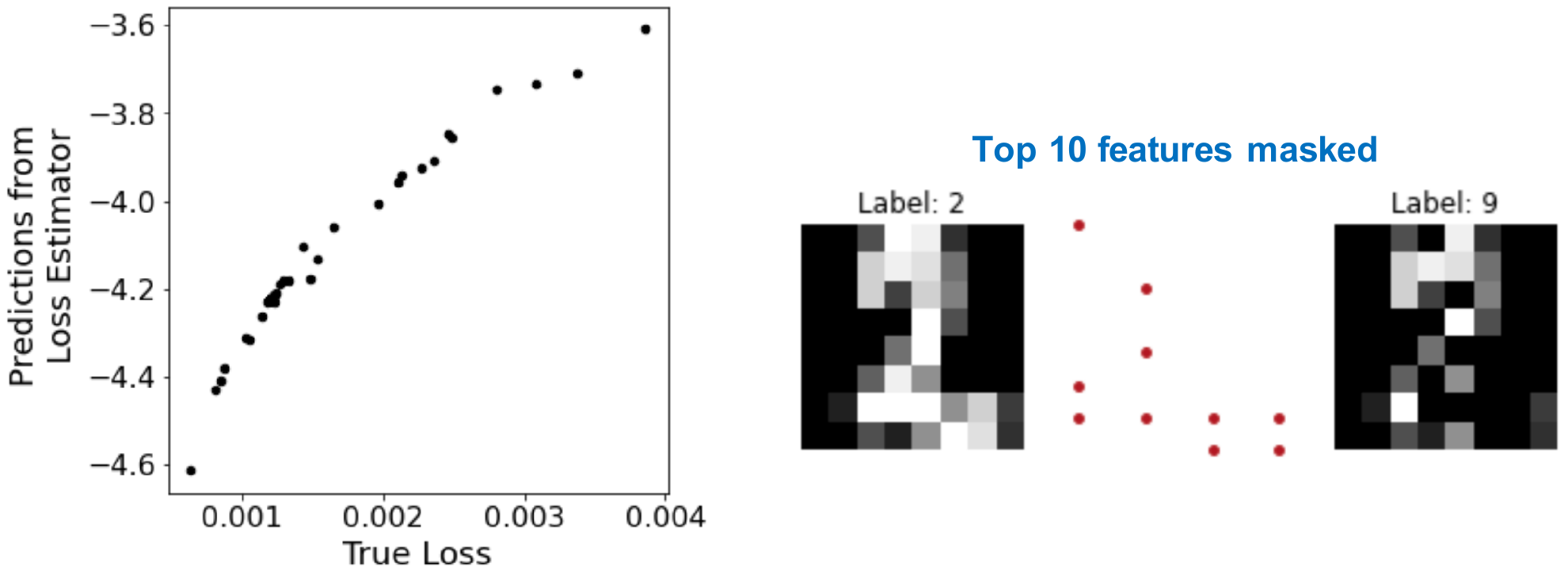}
    \caption{Demonstration of \name{} using the UCI handwritten digits dataset. Here, we show an example where the loss estimator was trained using the contrastive loss. For this test sample, the ranking obtained using the estimated loss agrees with that from the true loss (known ground truth). When we mask the top $10$ features from \name{} and as expected, there is a change in the model prediction.}
    \label{fig:demo}
\end{figure}

\begin{figure*}[t]
    \centering
    \subfloat[S][UCI Digits ($k = 15\%$)]{\includegraphics[width=0.33\linewidth,keepaspectratio]{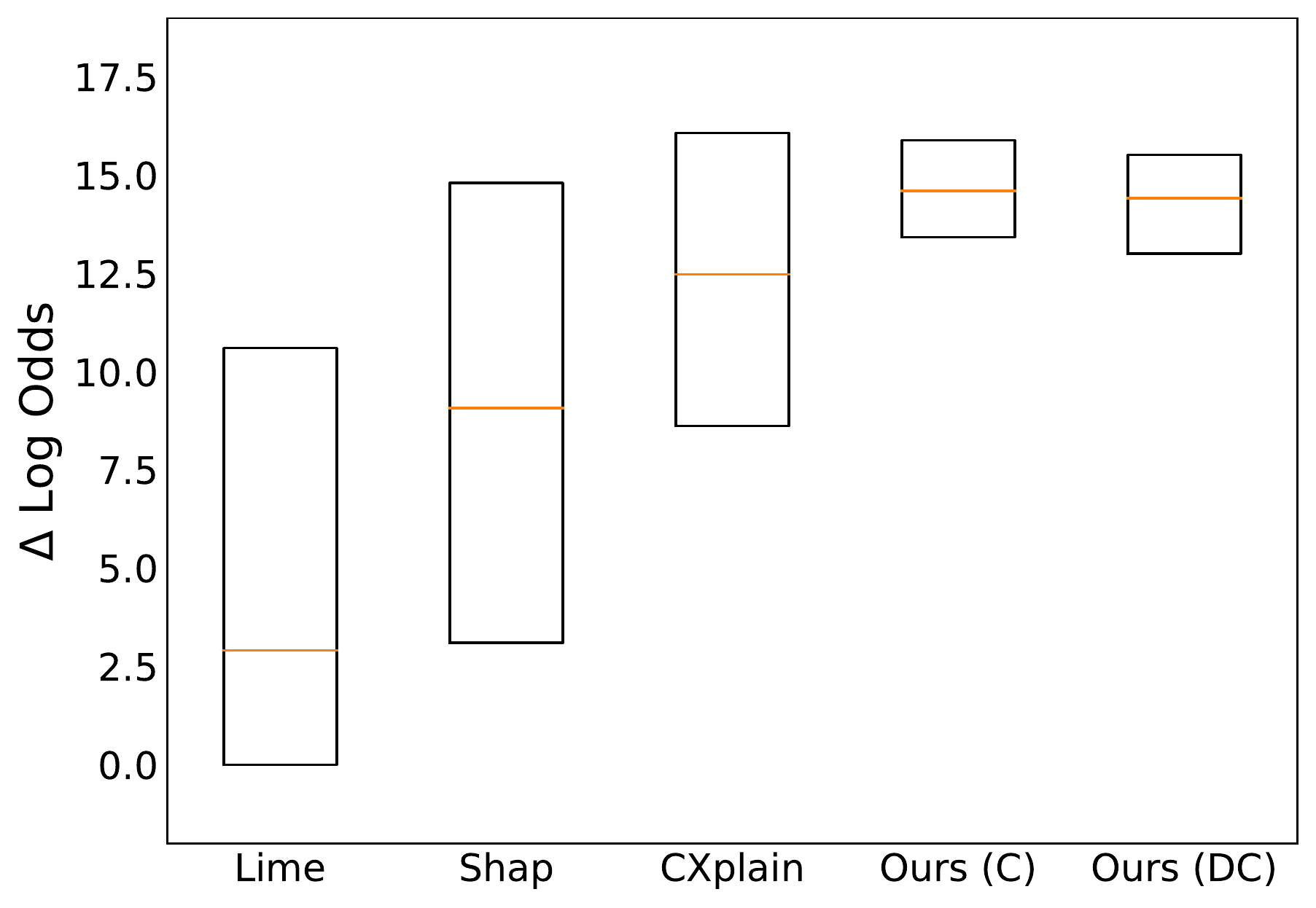}\label{fig:digits}}
    \subfloat[S][Kropt ($k = 15\%$)]{\includegraphics[width=0.33\linewidth,keepaspectratio]{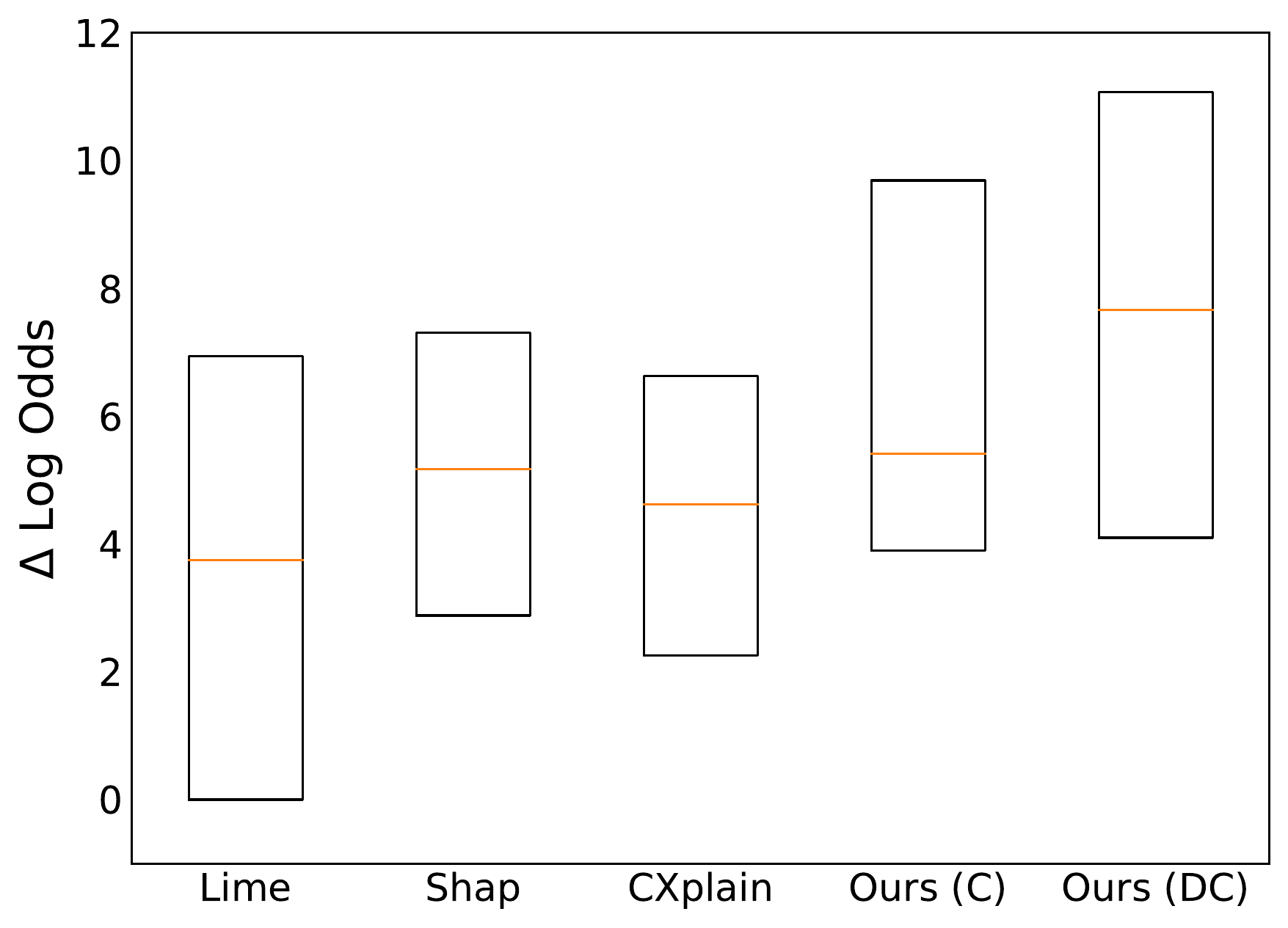}\label{fig:kropt}}
    \subfloat[S][RBF ($k = 15\%$)]{\includegraphics[width=0.33\linewidth,keepaspectratio]{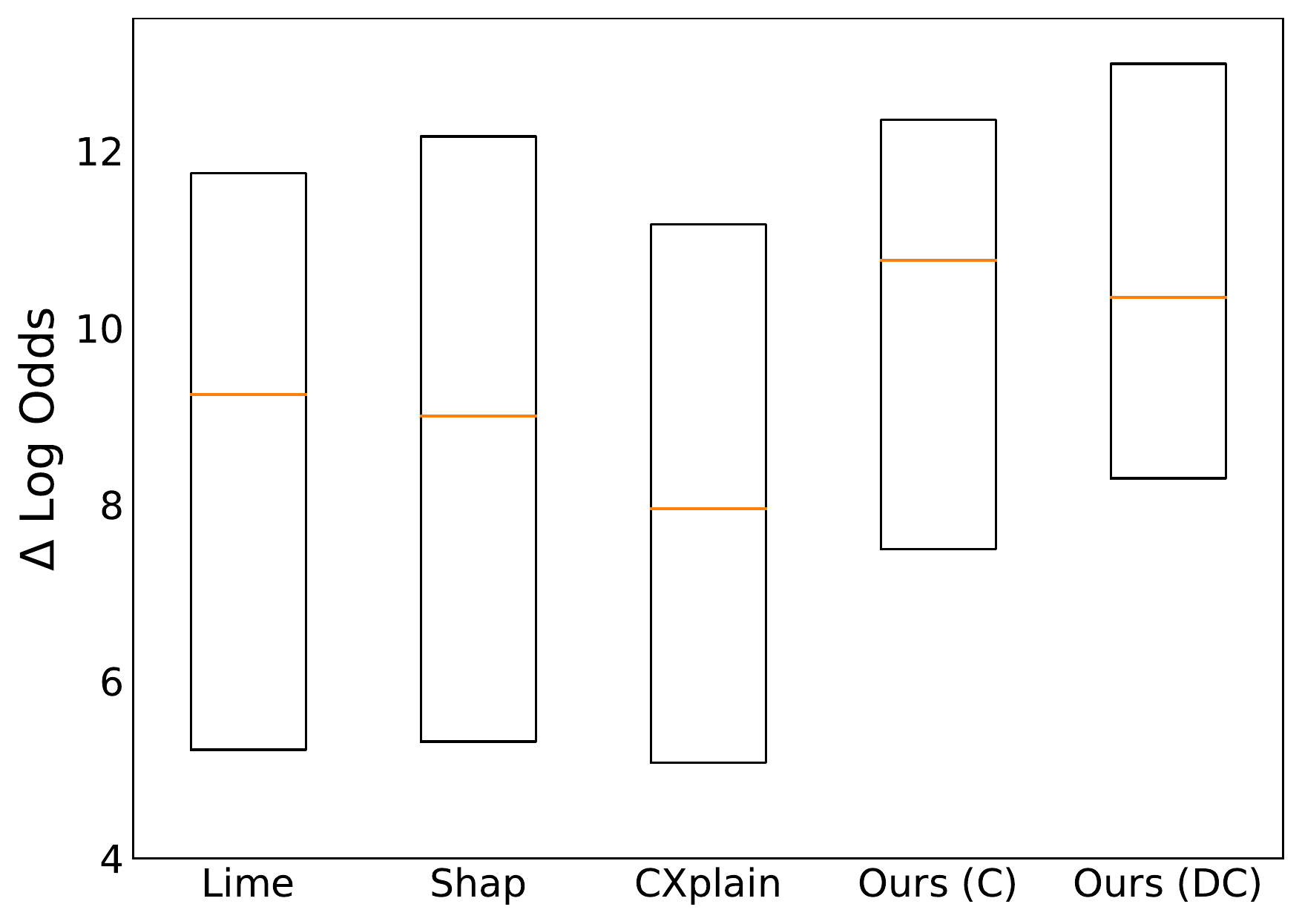}\label{fig:rbf}}
    \vfill
    \subfloat[S][Poker ($k = 15\%$)]{\includegraphics[width=0.33\linewidth,keepaspectratio]{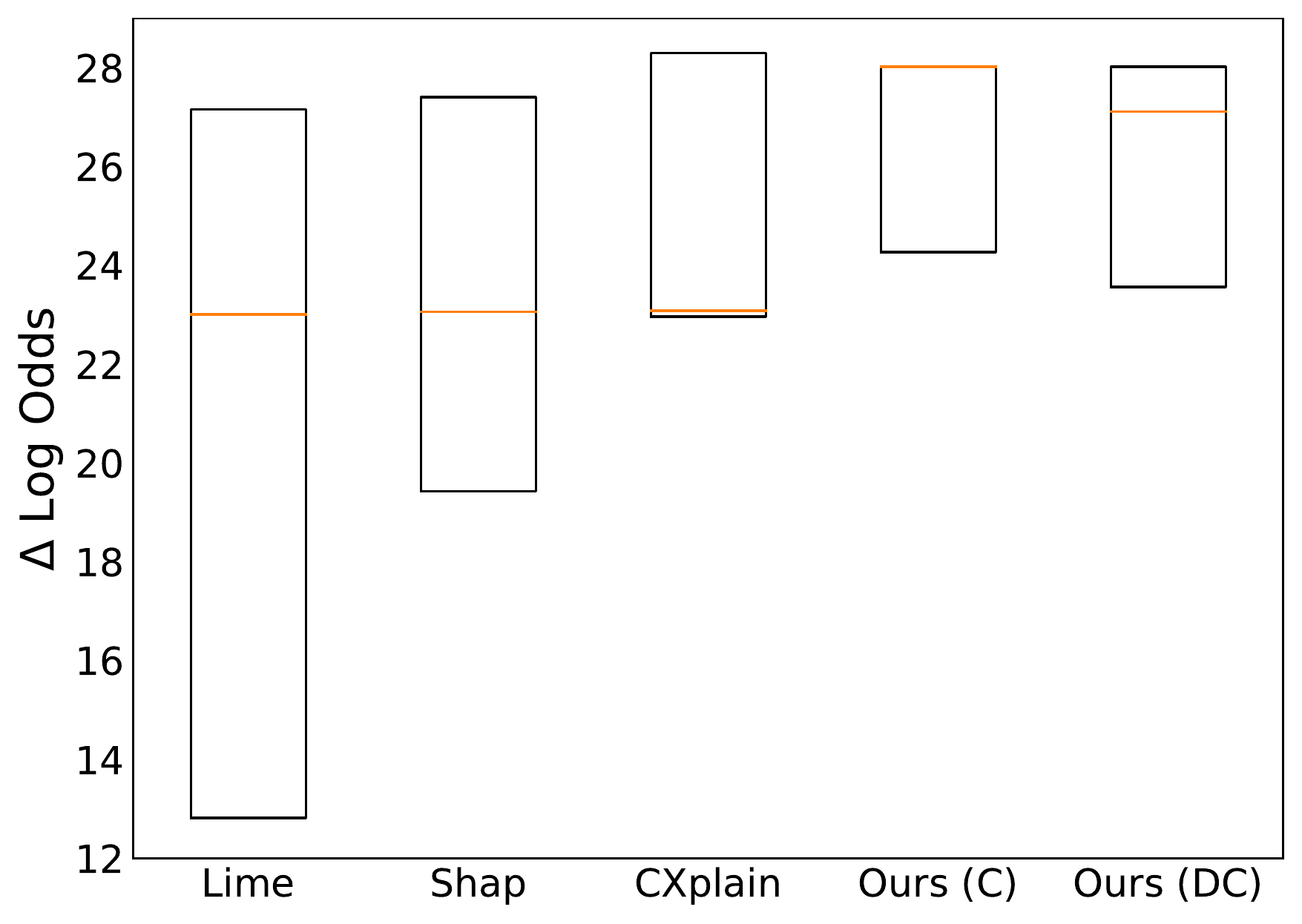}\label{fig:poker}}
    \subfloat[S][Handwritten Letters ($k = 15\%$)]{\includegraphics[width=0.33\linewidth,keepaspectratio]{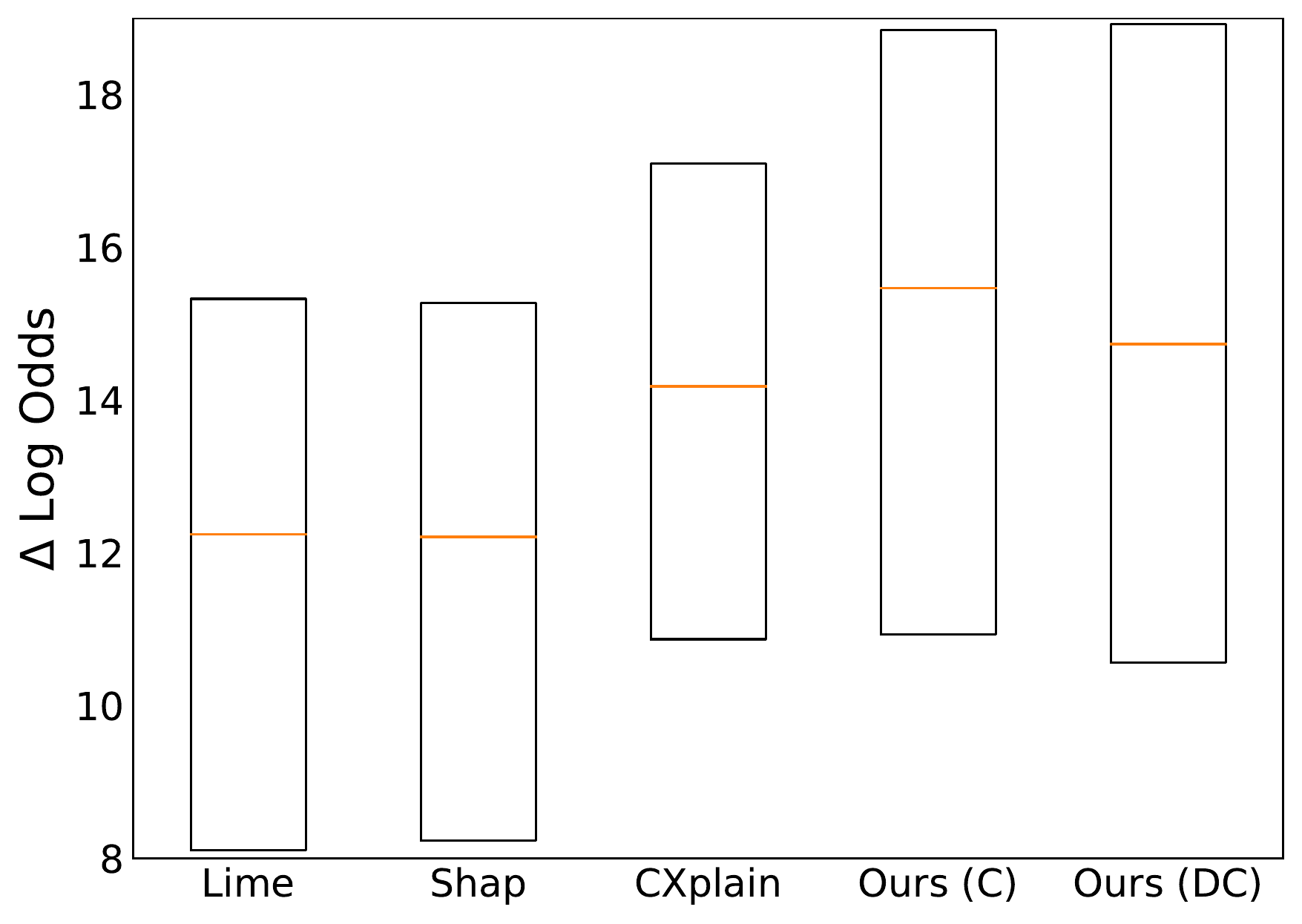}\label{fig:letter}}
    \subfloat[S][Cifar-10 ($k = 15\%$)]{\includegraphics[width=0.33\linewidth,keepaspectratio]{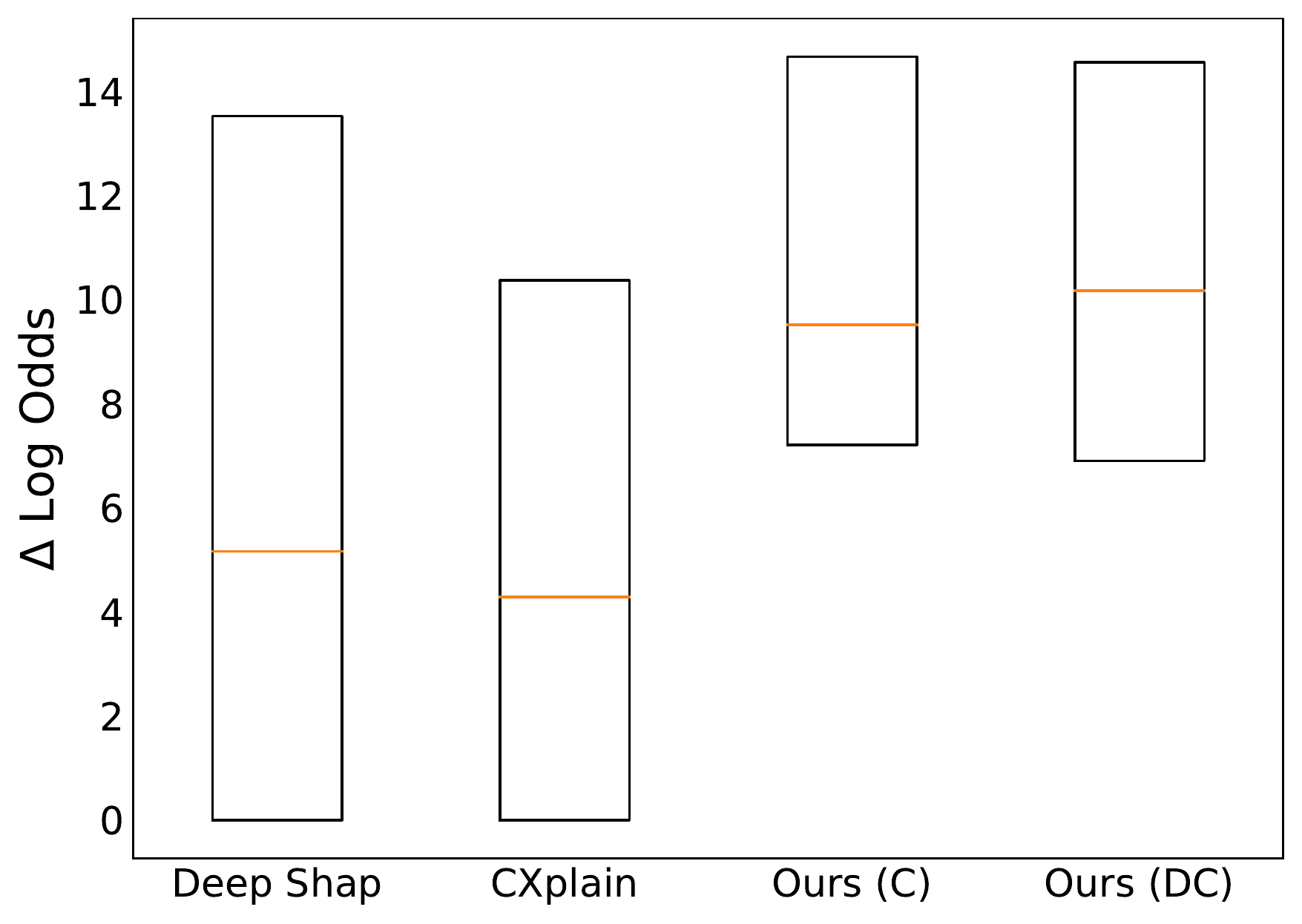}\label{fig:cifar-10}}
    \caption{Comparing the fidelity of feature importances inferred using different methods. We use the $\Delta$log-odds score (higher the better) obtained by masking the most influential input features. For each of the datasets, the ratio of features masked is also included in parentheses. Across all benchmarks, the proposed approach is consistently superior over the baselines.}
    \label{fig:perf}
\end{figure*}

\paragraph{Feature Importance Estimation.}Given the loss estimator $\mathcal{G}$, we estimate the feature importance using a Granger causality-based objective, similar to~\cite{schwab2019cxplain}. The Humean definition of causality adopted by Granger~\cite{granger1969investigating} postulates that a causal relationship exists between random variables $x_j$ and $y$, i.e., $x_j \rightarrow y$, if we can better predict using all available information than the case where the variable $x_j$ was excluded. This definition is directly applicable to our setting since it satisfies the key assumptions of Granger causality analysis, our data sample $\mathbf{x}$ contains all relevant variables required to predict the target and $\mathbf{x}$ temporally precedes $\mathbf{y}$. Mathematically,
\begin{equation}
    \Delta \epsilon_{\mathbf{x},j} = \epsilon_{\mathbf{x}\setminus \{j\}} - \epsilon_{\mathbf{x}},
    \label{eqn:causal}
\end{equation}where $\epsilon$ denotes the model error. For a sample $\mathbf{x}$, we can compute this objective for each feature $j$ to construct the explanation. As showed in Figure \ref{fig:arch}(bottom), we use the loss estimator to measure the predictive model's error in the presence and absence of a variable $x_j$ to check if $x_j$ causes the predicted output.
There are a variety of strategies that can be adopted to construct $\mathbf{x}\setminus \{j\}$. In the simplest form, we can mask the chosen feature by replacing it with zero or a pre-specified constant. However, in practice, one can also adopt more sophisticated masking strategies that take into account the underlying data distribution~\cite{janzing2013quantifying, vstrumbelj2009explaining}. Interestingly, our approach is agnostic to the masking strategy and the loss estimator can be used to compute the causal objective in Eqn.\eqref{eqn:causal} for any type of masking. In contrast, existing approaches such as CXPlain requires re-training of the explanation model for the new masking strategy.
% \ptb{I would add a sentence here mentioning that you could do the same thing for any pertubation based explanation since we do mention that in the introduction.}

Since the loss estimator is jointly trained with the main predictor, our approach does not require any additional adversarial training as done in~\cite{hima2020} to ensure that the explanations are consistent with the black-box. Furthermore, existing works in the active learning literature have found that the loss function~\cite{yoo2019learning} or its gradients~\cite{ash2019deep} effectively capture the inherent uncertainties in a model and hence can be used for selecting informative samples. Using a similar argument, we show that even though our causal objective is similar to CXPlain, our approach more effectively generalizes to even complex distribution shifts where CXPlain fails. 

\noindent \textbf{Demonstration.} For demonstrating the behavior of our approach, we consider the UCI handwritten digits dataset~\cite{dua2019} comprised of $8 \times 8$ grayscale images. In Figure \ref{fig:demo}, we show predictions from our loss estimator (contrastive training) for a test image, when each of the 64 pixels were masked (replaced with zero). We find that, though the scale of the loss function is discarded, the ordering of the features is well preserved. We also illustrate the explanation obtained by masking the top $10$ features identified using the causal objective in Eqn.\eqref{eqn:causal}. The observed changes in the prediction (from class $2$ to class $9$) is intuitive and demonstrates the effectiveness of our approach.

%%% Local Variables:
%%% mode: latex
%%% TeX-master: "main"
%%% End:

\section{Empirical Results}
% In this section, we present empirical studies to compare the proposed approach against popular baseline techniques for feature importance estimation using both non-image and image benchmark datasets. More importantly, we evaluate the fidelity of the inferred explanations under challenging distribution shifts and demonstrate the effectiveness of \name{}. Before we discuss our findings, we will discuss the datasets, metrics and baseline techniques used in our study.

In this section, we present empirical studies to compare \name{} against popular baselines using both non-image and image benchmark datasets. More importantly, we evaluate the fidelity of the inferred explanations under challenging distribution shifts and demonstrate the effectiveness of \name{}. Before we present our findings, we will discuss the datasets, baselines and metrics used in our study.

\begin{figure*}[t]
    \centering
    \includegraphics[width=0.99\linewidth,keepaspectratio]{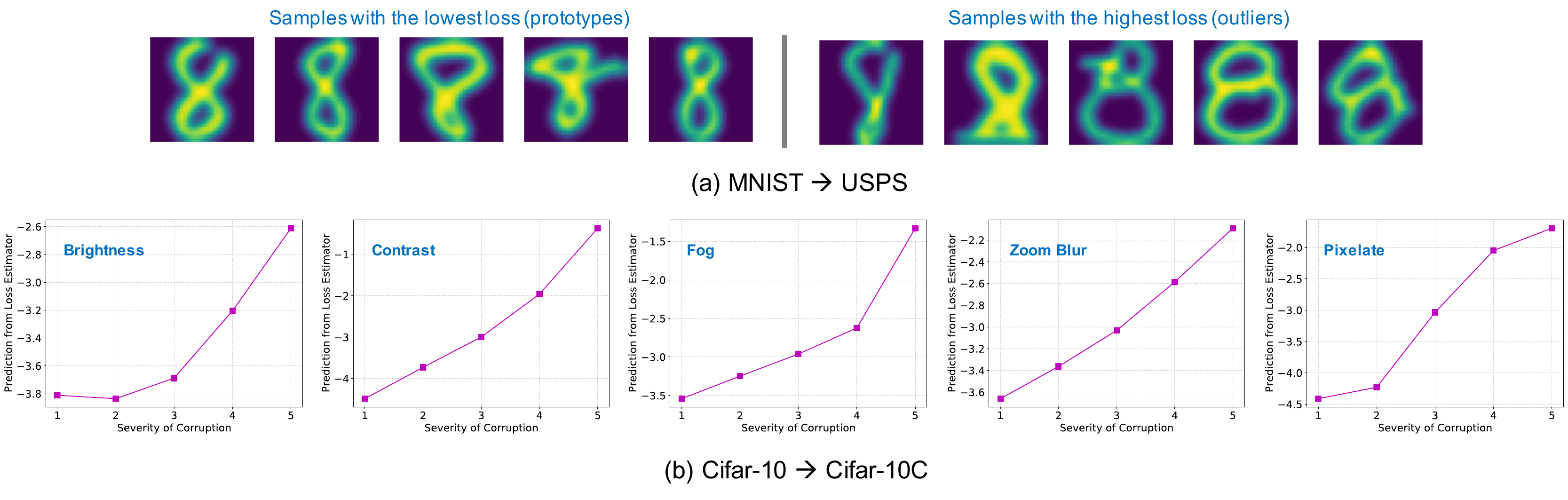}
    \caption{Effectiveness of our loss estimator $\mathcal{G}$ in detecting distribution shifts, even though the shifts are not known during training. In the MNIST-USPS case, it attributes non-typical writing styles from the USPS dataset, that are not found in MNIST, with high loss values. Similarly, in the case of CIFAR-10C, the loss estimates from $\mathcal{G}$, averaged across $500$ test samples, monotonically grows as the severity of the corruption increases.}
    \label{fig:distshift}
\end{figure*}

% \paragraph{Datasets.}We consider a suite of synthetic and real-world datasets to evaluate the fidelity and robustness of our approach. For the fidelity comparison study under standard testing conditions, we use:(a) UCI Handwritten Digits dataset~\cite{dua2019}, (b) OpenML benchmarks~\cite{OpenML2013}, namely Kropt, Letter Image Recognition, Pokerhand and RBF datasets, (c) Cifar-10 image classification dataset~\cite{krizhevsky2009learning}. The dimensionality of the input data ranged between $10$ and $64$, while the number of examples varied between $1797$ and $13750$. In each case, we utilized $\sim90\%$ of the data for training and the rest for evaluating the explanations.

\paragraph{Datasets.}We consider a suite of synthetic and real-world datasets to evaluate the fidelity and robustness of our approach. For the fidelity comparison study under standard testing conditions, we use the: (a) UCI Handwritten Digits dataset, (b) OpenML benchmarks~\cite{OpenML2013}, Kropt, Letter Image Recognition, Pokerhand and RBF datasets and (c) Cifar-10 image classification dataset~\cite{krizhevsky2009learning}. The dimensionality of the input data ranges from $10$ to $64$ and the total number of examples varies between $1797$ and $13750$ for different benchmarks. For each of the UCI and OpenML datasets, we utilized $\sim90\%$ of the data while for Cifar-10, we used the prescribed dataset of $50$K RGB images of size $32$$\times$$32$ for training our proposed model. 

For the robustness study, we used the following datasets: (a) \textit{Synthetic dataset}: In order to study the impact of distribution shifts on explanation fidelity, we constructed synthetic data based on correlation and variance shifts to the data generation process defined using a multi-variate normal distribution. More specifically, we generated multiple synthetic datasets of $5$K samples, where the number of covariates was randomly varied between $10$ and $50$. In each case, the samples were drawn from $\mathcal{N}(\boldsymbol{\mu}, \mathbf{\Sigma})$, where $\mu_{ii} = \alpha, \Sigma_{ii} = 1$ and $\Sigma_{ij} = \beta$ and the values $\alpha$, $\beta$ (the correlation between any two variables) were randomly chosen from the uniform intervals $[-2,2]$ and $[-1,1]$ respectively. The label for each sample was generated using their corresponding quantiles (i.e., defining classes separated by nested concentric multi-dimensional spheres). To generate correlation shifts, we created new datasets following the same procedure, but using a different correlation $\bar{\beta} = \beta + \delta_{\beta}$. Here, $\delta_{\beta}$ was randomly drawn from the uniform interval $[-0.2,0.2]$. Next, we created a third dataset to emulate variance shifts, wherein we changed the variance $\Sigma_{ii} = \Sigma_{ii} + \kappa$ and $\kappa$ was drawn from the uniform interval $[0.25,0.75]$. While the predictive model was trained only using the original dataset, the explanations were evaluated using both correlation- and variance-shifted datasets. We generated $10$ different realizations with this process and report the explanation fidelity metrics averaged across the $10$ trials; (b) \textit{Cifar10 to Cifar10-C}~\cite{hendrycks2019robustness}: This is a popular benchmark for distribution-shift studies, wherein we train the predictive model and loss estimator using the standard Cifar10 dataset and generate explanations for images from the Cifar10-C dataset containing wide-variety of natural image corruptions; and (c) \textit{MNIST-USPS}: In this case, we train the predictive model using only the MNIST handwritten digits dataset~\cite{lecun2010mnist} and evaluate the explanations on the USPS dataset~\cite{usps} at test time. 
\begin{figure}[t]
    \centering
    \includegraphics[width=0.5\linewidth,keepaspectratio]{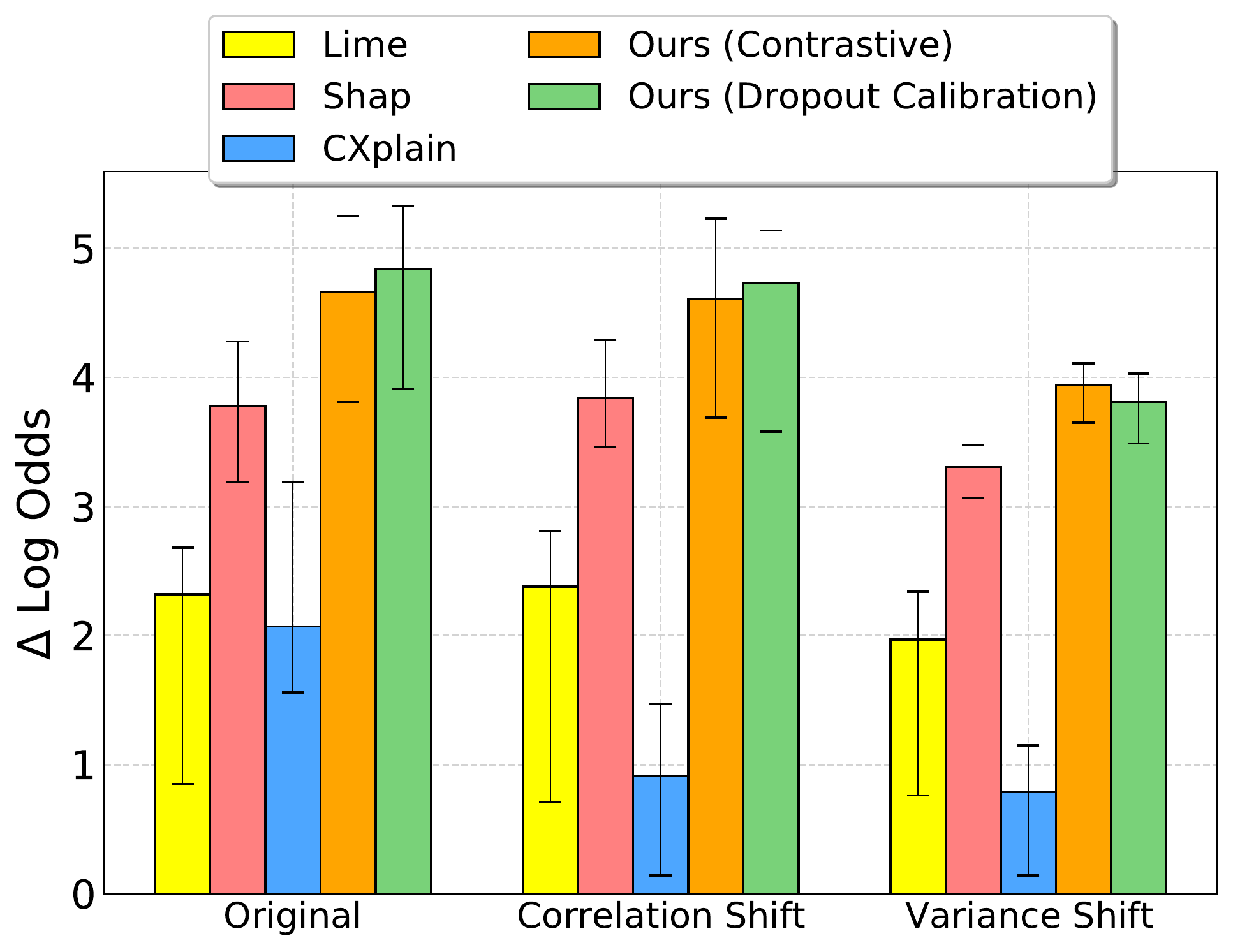}
    \caption{Using a synthetic dataset to study the robustness of explanations obtained using different approaches, under correlation and variance shifts. We mask the top $25$\% of features in the data to obtain the $\Delta$log-odds scores.}
    \label{fig:synthetic}
\end{figure}

\begin{figure*}[t]
    \centering
    \includegraphics[width=0.95\linewidth,keepaspectratio]{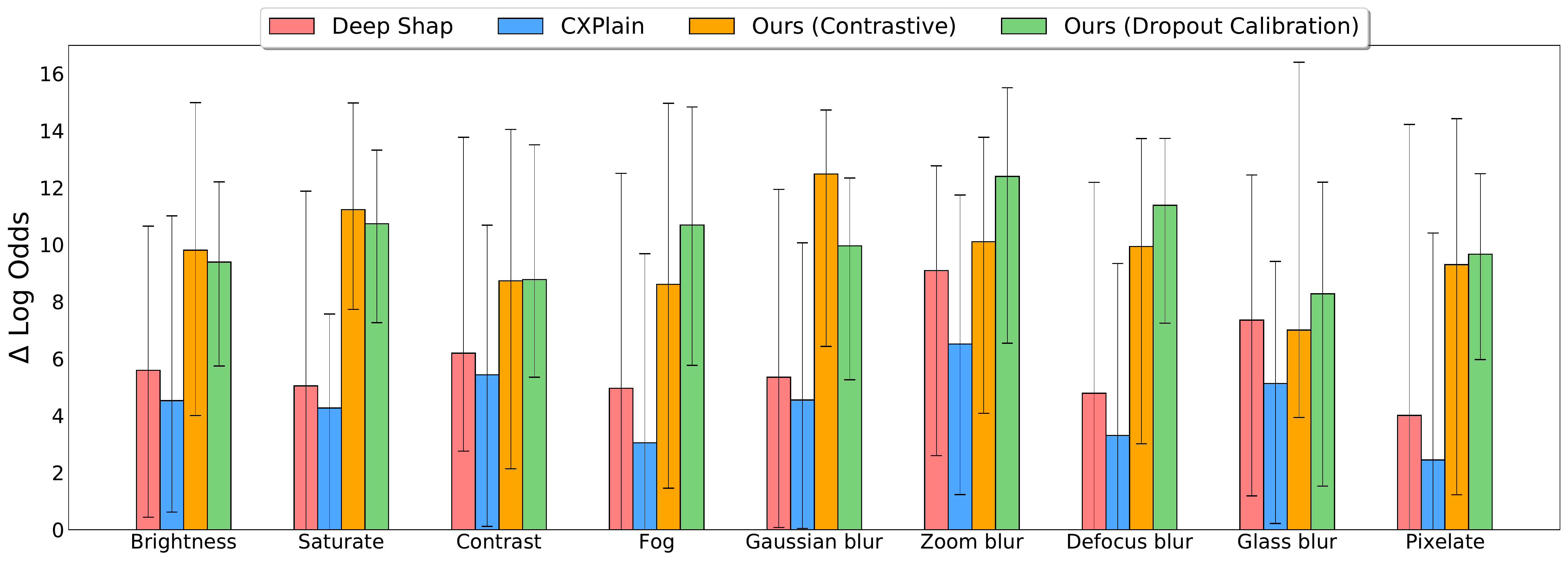}
    \caption{\textit{Cifar-10C dataset}: We study the fidelity of explanations generated on different types of corrupted images using the loss estimator trained on the original Cifar-10 data.}
    \label{fig:cifar-ex}
\end{figure*}

\paragraph{Baselines.}We compared~\name{}~ against the following baseline methods that are commonly adopted to produce sample-level explanations. All baseline methods considered belong to the class of post-hoc explanation strategies which aim to construct interpretable models that can approximate the functionality of any black-box predictor.

\noindent \emph{(i) LIME}~\cite{ribeiro2016}\footnote{code: \url{https://github.com/marcotcr/lime}}: LIME constructs linear models, which can locally approximate a black box predictor, by fitting a weighted regression model around the sample to be explained based on variants of the sample obtained by perturbing or zero-masking the input features. The intuition is that the post-hoc regression model obtained is reflective of the sensitivity of the black-box predictor to the modifications in the input features. The coefficients of the obtained post-hoc model serve as attribution scores for each feature in the given sample. 

\noindent \emph{(ii) Shap}~\cite{lundberg2017unified}\footnote{code: \url{https://github.com/slundberg/shap}}: SHAP determines the feature attribution scores for a sample by marginalizing the individual contributions of every feature towards a prediction. SHAP, more specifically KernelSHAP, fits a local regression model around the sample to be explained using multiple realizations of the sample by zero masking single or groups of features. A fundamental difference between LIME and SHAP lies in the SHAP kernel used, which is a function of the cardinality of the features present in a group. The coefficients of the obtained model are the SHAPLey attribution scores for every feature in the given sample. 

\noindent \emph{(iii) CXPlain}~\cite{schwab2019cxplain}\footnote{code: \url{https://github.com/d909b/cxplain}}:  This determines feature attribution scores by training a post-hoc model that learns to approximate the distribution of Granger causal errors~\cite{granger1969investigating}, i.e., the difference between the black-box prediction loss when no feature is masked and the loss when features are zero-masked one at a time. The feature attribution scores obtained from the model are thus reflective of the global distribution of the causality based error metric. Similar to~\cite{schwab2019cxplain}, we use an MLP and a U-Net model as the post-hoc explainer for the non-image and the image datasets respectively in our experiments.  

\noindent \emph{(iv) Deep Shap}~\cite{lundberg2017unified} DeepSHAP is a fast and scalable approximation of SHAP and also closely related to the DeepLIFT algorithm. We utilize this baseline on datasets where LIME and SHAP were expensive to run.

% \paragraph{Evaluation Metric.} To evaluate the explanation fidelity, we utilize the commonly used \textit{difference in log-odds} metric, which is a measure of change in prediction when $k\%$ of the most relevant features in the input data are masked.
% \begin{equation}
%     \Delta \text{log-odds} = \text{log-odds}(\text{p}_{\text{ref}}) - \text{log-odds}(\text{p}_{\text{masked}}),
% \end{equation}where $\text{log-odds}(\text{p}) = \text{log}(\frac{\text{p}}{1-\text{p}})$ and $\text{p}_{\text{ref}}$ is the reference prediction probability of the original data and $\text{p}_{\text{masked}}$ refers to the prediction probability when a subset of features are masked. A higher value for $\Delta \text{log-odds}$ implies higher fidelity of the feature importance estimation.

\paragraph{Evaluation Metric.} To evaluate the explanation fidelity, we utilize the commonly used difference in log-odds metric, which is a measure of change in prediction when $k\%$ of the most relevant features in the input data are masked.
\begin{equation}
    \Delta \text{log-odds} = \text{log-odds}(\text{p}_{\text{ref}}) - \text{log-odds}(\text{p}_{\text{masked}})
\end{equation}
Here $\text{log-odds}(\text{p}) = \text{log}(\frac{\text{p}}{1-\text{p}})$ and $\text{p}_{\text{ref}}$ is the reference prediction probability of the original data and $\text{p}_{\text{masked}}$ refers to the prediction probability when a subset of features are masked. A higher value for $\Delta \text{log-odds}$ implies higher fidelity of the feature importance estimation. More specifically, for: (a) \textit{Non-Image Datasets}. We sort the feature attribution scores obtained from the explainability method (\name{} and baselines) and zero mask the top $k\%$ important features in the input sample to evaluate the metric, and (b) \textit{Image Datasets}. We use the SLIC~\cite{achanta2012slic} segmentation algorithm to generate superpixels, which are then used to compute the feature importance scores. For CXPlain and DeepSHAP, we aggregate the pixel-level feature importance scores to estimate attributions for each superpixel. 
\begin{figure*}[!t]
    \centering
    \includegraphics[width=0.95\linewidth,keepaspectratio]{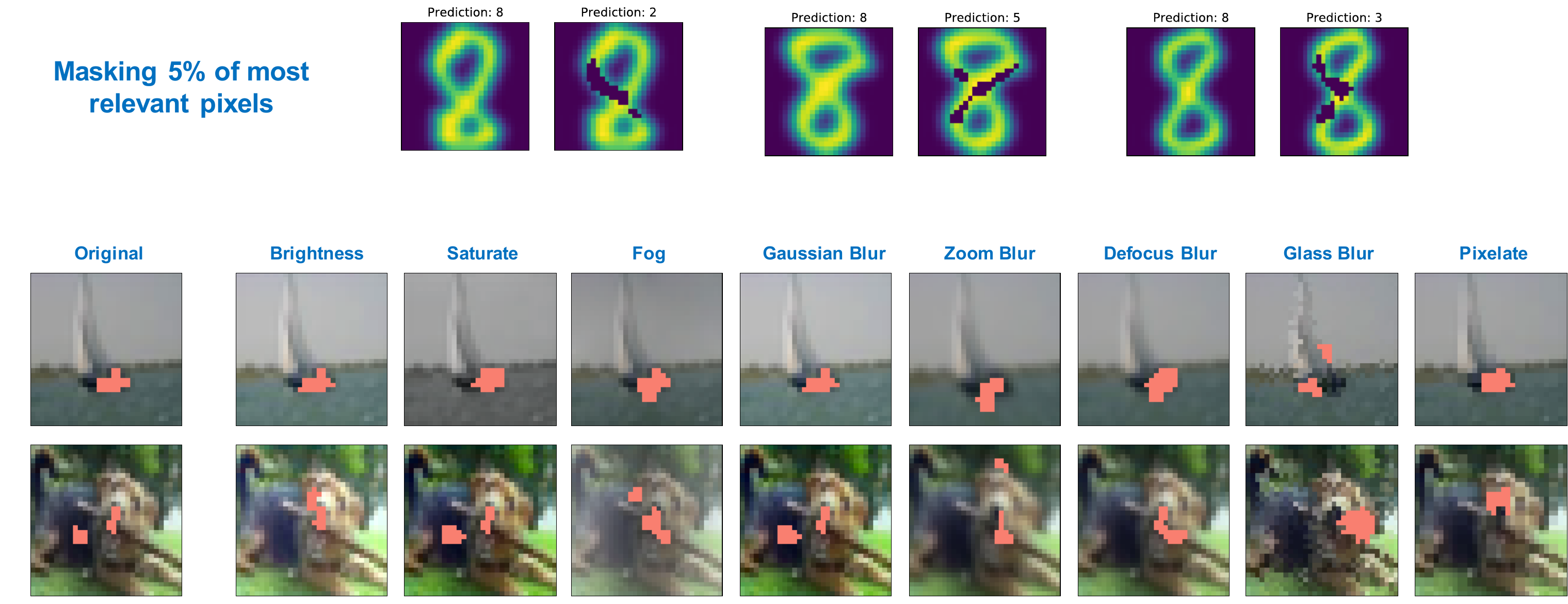}
    \caption{Examples of explanations generated using the proposed approach (with dropout calibration) on USPS and Cifar-10C datasets using models trained with MNIST and Cifar-10 respectively.}
    \label{fig:exp}
\end{figure*}

\paragraph{Hyperparameters.}For all non-imaging datasets, the black-box model was a 5 layer MLP with ReLU activations, each fully-connected (FC) layer in the loss estimator contained $16$ units. In the case of Cifar-10, we used the standard ResNet-18 architecture, and the loss estimator used outputs from each residual blocks (with fully connected layers containing $128$ hidden units). Finally, for the MNIST-USPS experiment, we used a $3-$layer CNN with $2$ FC layers. The loss estimator was designed to access outputs from the first $4$ layers of the network and utilized FC layers with $16$ units each. All networks were trained using the ADAM optimizer with a learning rate $0.001$ and batch size $128$.

\subsection{Findings}
\subsubsection{PRoFILE Produces Higher Fidelity Explanations}
Figure \ref{fig:perf} illustrates the $\Delta \text{log-odds}$ obtained using PRoFILE with both the proposed learning strategies (Ours(C) and Ours(DC)) in comparison to the baselines. Note that, for the UCI and OpenML datasets, we used the held-out test set for our evaluation ($90$-$10$ split), while for Cifar-10, we used $50$ randomly chosen test images for computing the fidelity metric. While \name{} and CXPlain are scalable to larger test sizes, the small subset of test samples was used to tractably run the other baselines. For each dataset, we show the median (orange), along with the ${25}^{th}$ and the ${75}^{th}$ percentiles, of the $\Delta \text{log-odds}$ scores across the test samples. We find that PRoFILE consistently outperforms the existing baselines on all benchmarks. In particular, both contrastive training and dropout calibration strategies are effective and perform similarly in all cases. The improved fidelity can be attributed directly to the efficacy of the loss estimator and the causal objective used for inferring the feature attribution. In comparison, both LIME and SHAP produce lower fidelity explanations, while also being computationally inefficient. Interestingly, though CXPlain also uses a causal objective similar to us, the resulting explanations are of significantly lower fidelity. In terms of computational complexity for generating post-hoc explanations, \name{} which requires $p$ evaluations (number of features that need to be masked and can be parallelized) of the loss estimator and is only marginally more expensive than CXPlain.

\subsubsection{Loss Estimator Detects Distribution Shifts}
At the core of our approach is the pre-trained loss estimator, which enables us to utilize the Granger causality objective to generate high-quality explanations. Consequently, the robustness of PRoFILE directly relies on how well the loss estimator can generalize under distribution shifts. We investigate the empirical behavior of the loss estimator using (i) MNIST-USPS and (ii) Cifar10 to Cifar10-C benchmarks. In both cases, we train the predictor and loss estimator using the original data (MNIST, Cifar10) and evaluate on the shifted data. In Figure \ref{fig:distshift}(a), we show USPS images from class $8$ with the lowest (in-distribution) and highest (out-distribution) loss estimates. While the former resemble the prototypical examples, the latter contains uncommon writing styles not found in the MNIST dataset. In case of Cifar10-C, we show the loss estimates for $5$ different natural image corruptions (averaged across $500$ examples). We observe a monotonic increase in the average loss estimates as the severity of the corruptions grow, thus demonstrating the ability of the loss estimator to detect distribution shifts. 

%  Our models are trained only on either MNIST or Cifar-10 while the evaluation is on the USPS or Cifar-10-C data respectively. It can be observed  that the loss estimator produces low losses for images (Digit 8 in this example) that resembles the prototypes in the MNIST data while higher loss values signify uncommon images of the digit 8 found in the MNIST dataset. 

\subsubsection{PRoFILE Explanations are More Robust}
Following our observations on the behavior of the loss estimator, we now evaluate the fidelity of PRoFILE explanations in those scenarios. Figure \ref{fig:synthetic} illustrates the median $\Delta \text{log-odds}$ and error bars obtained by masking the top $25$\% of features on $10$ realizations of the synthetic dataset. In particular, we show the results for the held-out correlation and variance shifted data, while the models were trained only using the original synthetic data. We find that by utilizing a pre-trained loss estimator, \name{} significantly outperforms the baselines, even under complex shifts, indicating the robustness of our approach. Similar to the findings in~\cite{hima2020}, we note that the widely-adopted baselines are not immune to shifts. Figure \ref{fig:cifar-ex} shows a detailed comparison of $\Delta \text{log-odds}$ for the Cifar10-C dataset. Note, we show the median, $25^{\text{th}}$ and $75^{\text{th}}$ percentiles. We find that PRoFILE consistently achieves superior fidelity, when compared to existing baselines, except in the case of \textit{glass blur} where the scores are comparable.

Figure \ref{fig:exp} shows examples of explanations obtained using PRoFILE on the USPS and Cifar10-C datasets. We can observe from Figure \ref{fig:exp} (top) that our method adapts well across domains to identify critical pixels that characterize class-specific decision regions. Interestingly, these are examples where digit $8$ is suitably masked by PRoFILE (only $5\%$ of pixels) to be predicted as one of the other classes sharing the decision boundary. It can also be observed from Figure \ref{fig:exp}(bottom) that the PRoFILE explanations obtained under different domain shifts are consistent. We note that, in all cases except~\textit{glass blur}, it identifies the hull of the boat and the mouth of the dog as critical features. These observations strongly corroborate with the performance improvements in Figure~\ref{fig:cifar-ex}.

\section{Conclusions}
In this paper, we proposed \name{}, a novel post-hoc feature importance estimation method applicable to any data modality or architecture. In particular, \name{} trains an auxiliary estimator to estimate the expected loss, for a given sample, from the primary predictor model. To this end, we introduced two learning objectives, contrastive training and dropout calibration. Using the pre-trained loss estimator along with a causality based objective, \name{} can accurately estimate feature importance scores that are immune to a wide variety of distribution shifts. Through extensive experimental studies on different data modalities, we demonstrate that \name{} prvoides higher fidelity expalantions, is robust under real-world distribution shifts and is computational effective when compared to commonly adopted feature importance estimation methods.

 \section{Acknowledgements}
This work was performed under the auspices of the U.S. Department of Energy by
Lawrence Livermore National Laboratory under Contract DE-AC52-07NA27344.

\bibliographystyle{aaai21}
\bibliography{refs}

\end{document}